\documentclass[letterpaper, 10 pt, conference]{ieeeconf}  

\IEEEoverridecommandlockouts                              

\makeatletter
\let\NAT@parse\undefined
\makeatother
\usepackage[numbers]{natbib}

\renewcommand{\baselinestretch}{0.96}

\usepackage{xcolor}

\usepackage{float}
\usepackage{graphics}
\usepackage{graphicx}
\usepackage{multirow}
\usepackage{amsmath}

\begin{document}

\title{\LARGE \bf
UniGrasp: Learning a Unified Model to Grasp with\\ Multifingered Robotic Hands}

\author{Lin Shao$^{1}$, Fabio Ferreira$^{*1,2}$, Mikael Jorda$^{*1}$, Varun Nambiar$^{*1}$, Jianlan Luo$^{3}$, Eugen Solowjow$^{4}$,\\  Juan Aparicio Ojea$^{4}$, Oussama Khatib$^{1}$, Jeannette Bohg$^{1}$
\thanks{This work has been partially supported by JD.com American Technologies Corporation (“JD”) under the SAIL-JD AI Research Initiative and by the International Center for Advanced Communication Technologies (InterACT). This article solely reflects the opinions and conclusions of its authors and not JD or any entity associated with JD.com.}
\thanks{$^*$The authors contributed equally.}
\thanks{Lin Shao, Fabio Ferreira, Mikael Jorda, Varun Nambiar, Oussama Khatib, Jeannette Bohg are with Stanford Artificial Intelligence Lab (SAIL), Stanford University, Stanford, CA, USA. {\tt\footnotesize [lins2,fabiof,mjorda,vnambiar,khatib,bohg]
@stanford.edu}}%
\thanks{Fabio Ferreira is with Institute for Anthropomatics and Robotics, Karlsruhe Institute of Technology, Karlsruhe, Germany. {\tt\footnotesize fabioferreira@mailbox.org}}%
\thanks{Jianlan Luo is with Dep. of ME and Dep. of EECS, University of California, Berkeley, CA, USA.  {\tt\footnotesize jianlanluo@cs.berkeley.edu}}
\thanks{Eugen Solowjow, Juan Aparicio Ojea are with Siemens  Corporate  Technology, Berkeley, CA, USA. {\tt\footnotesize [eugen.solowjow,
juan.aparicio]@siemens.com}}%
}


\maketitle

\begin{abstract}\label{sec:abstract}
To achieve a successful grasp, gripper attributes such as its geometry and kinematics play a role as important as the object geometry. The majority of previous work has focused on developing grasp methods that generalize over novel object geometry but are specific to a certain robot hand. We propose {\em UniGrasp\/}, an efficient data-driven grasp synthesis method that considers both the object geometry and gripper attributes as inputs. {\em UniGrasp\/} is based on a novel deep neural network architecture that selects sets of contact points from the input point cloud of the object. The proposed model is trained on a large dataset to produce contact points that are in force closure and reachable by the robot hand. By using contact points as output, we can transfer between a diverse set of multifingered robotic hands. Our model produces over 90\% valid contact points in Top10 predictions in simulation and more than 90\% successful grasps in real world experiments for various known two-fingered and three-fingered grippers. Our model also achieves 93\%, 83\% and 90\% successful grasps in real world experiments for an unseen two-fingered gripper and two unseen multi-fingered anthropomorphic robotic hands.
\end{abstract}


\section{Introduction}\label{sec:intro}
The ability of a robot to grasp a variety of objects is of tremendous importance for many application domains such as manufacturing or warehouse logistics.
It remains a challenging problem to find suitable grasps for arbitrary objects and grippers; especially when objects are only partially observed through noisy sensors. 

The most successful approaches in recent literature are data-driven methods where large-capacity models are trained on labeled training data to predict suitable grasp poses, e.g.~\cite{bohg2014,sahbani2013,lenz2015deep, kappler2015leveraging,pinto2016supersizing,mahler2017dex,mahler2019learning,arm-farm-first,kalashnikov18a,varley2015,schmidt2018grasping,veres2017modeling,lu2020planning}. The main objective of these types of approaches is to generalize to objects that were not part of the training data. All of these models are trained for one particular end-effector, which is typically a two-finger gripper. There are only few exceptions that consider more dexterous hands e.\,g.~\cite{kappler2015leveraging,varley2015,schmidt2018grasping,veres2017modeling,liang2019pointnetgpd,lu2020planning}. 
Of those, \cite{varley2015,veres2017modeling,lu2020planning} go beyond a small set of pre-grasp shapes. \cite{mahler2019learning} is the only approach we are aware of that considers two different grippers in one model but still trains separate grasp synthesis models for each one.

\begin{figure}[tb!]
 \centering
 \includegraphics[width=0.85\linewidth ]{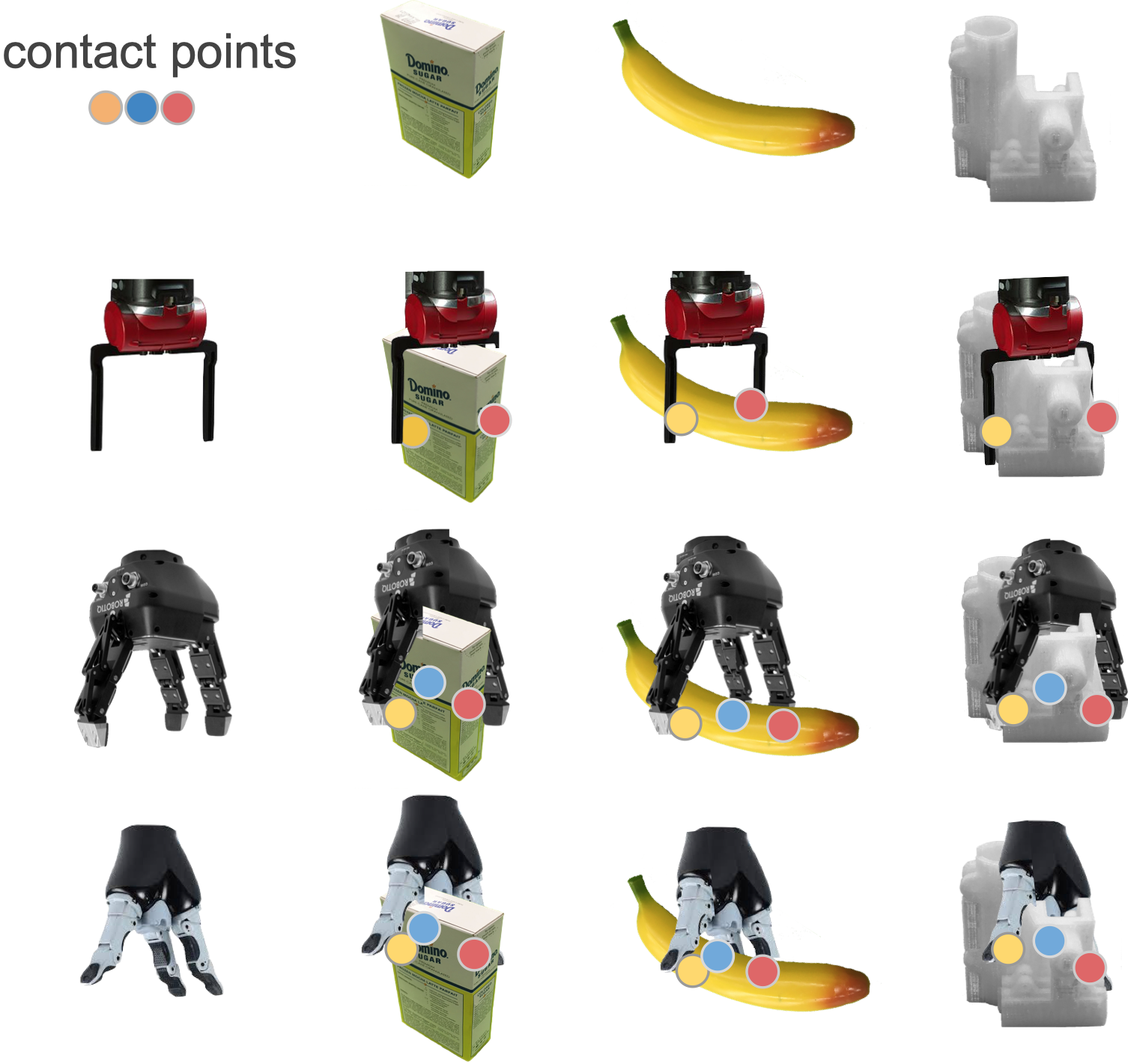}
 \caption{{\em UniGrasp\/} takes as input a kinematic description of a robotic hand and a point cloud of an object. Given this input, {\em UniGrasp\/} is trained on a large dataset to produce contact points on the object surface that are in force closure and reachable by the robotic hand. {\em UniGrasp\/} produces valid contact point sets not only on novel objects but also generalizes to new multifingered hands that it has not been trained on.}
\label{fig:teaser}
\end{figure}

In this paper, we go beyond generalization to unseen objects. We propose {\em UniGrasp\/}, a model that additionally generalizes to novel gripper kinematics and geometries with more than two fingers (see Fig.~\ref{fig:teaser}). Such a model enables a robot to cope with multiple, interchangeable or updated grippers without requiring retraining. 
{\em UniGrasp\/} takes as input a point cloud of the object and a robot hand specification in the {\em Unified Robot Description Format\/} (URDF)~\cite{urdf}. It outputs a set of contact points that are in force closure and reachable by the robot hand. This contact-based grasp representation allows to infer precision grasps that exploit the dexterity of the multi-fingered hand. It also alleviates the need to define a finite set of grasp pre-shapes and approach directions.

This work makes the following contributions: (1) Given a point cloud from an object and a URDF model of a gripper, we map gripper and object features into separate lower-dimensional latent spaces. The resulting features are concatenated and form the input to the part of the model that generates sets of contact points.
(2) We propose a novel, multi-stage model that selects sets of contact points from an object point cloud. These contact points are in force closure and reachable by the gripper to produce a 6D grasp.
(3) In simulation and the real world, we show that both, our mapping scheme and multi-stage model adapts well to a varying number of robot hands. The selected contact point sets are typically clustered in a local neighborhood. Such an attribute is well suited for execution on a real robot because precisely aligning the robotic fingers with an isolated point may be hard to achieve. 
(4) Finally, we publish a new large-scale dataset comprised of tuples of annotated objects with contact points for various grippers, covering many popular commercial grippers.

\section{Related Work}\label{sec:relatedwork}
In the following, we review related work that shares some of the key assumptions with our work, e.g. the object to be grasped is novel and only observed through a noisy depth camera from a single point of view.  There are grasp synthesis methods for multi-fingered hands~\cite{hang2014hierarchical,li2016dexterous,fan2018real} that do not rely on learning approaches. Different from our approach, they typically rely on full observations of the objects and take several seconds of computation. For a broader review of the field on data-driven grasp synthesis, we refer to~\cite{bohg2014}.  
Furthermore, we briefly discuss learning-based methods for processing point clouds.

\subsection{Data-Driven Grasp Synthesis for Novel Objects}
Learning to grasp novel objects from visual data has been an active field of research since the seminal paper by ~\citet{saxena2008robotic}. Since then, many approaches have been developed that use more sophisticated function approximators (CNNs) to map from input data to a good grasp~\cite{lenz2015deep, kappler2015leveraging,pinto2016supersizing,mahler2017dex,mahler2019learning,arm-farm-first,kalashnikov18a,varley2015,schmidt2018grasping,morrison2018closing,liang2019pointnetgpd,le2010learning}. These approaches differ in the input data format (e.g. 2D images, depth maps or point clouds), the grasp representation (e.g. grasping points, oriented 2D rectangles, a 2D position and wrist orientation or a full 6D pose) and whether they learn to grasp from real world examples or synthetic data. 

In our work, we take as input a point cloud of a segmented object and generate a set of contact points, one per finger of the hand. We train our model on synthetically generated data. The key difference to previous work is that we train one model that generalizes over many robot hands, even if they were not part of the training data. 
All of the aforementioned prior work is trained for one particular hand. Typically, this is a two-fingered gripper with exceptions including~\cite{kappler2015leveraging,varley2015,liang2019pointnetgpd,schmidt2018grasping,veres2017modeling,lu2020planning}. \cite{mahler2019learning} is the only approach we are aware of that considers two different grippers (a suction cup and a parallel jaw gripper). However, the authors still train separate grasp synthesis models for each gripper.
Another key difference to previous work is that we adopt contact points to represent a grasp. Contact points are rarely used in recent work because it is difficult to precisely make contact with these points under noise in perception and control~\cite{bohg2014}. For this reason, the dominant grasp representation in related work has been the pose of the end-effector, e.g.~\cite{kappler2015leveraging,arm-farm-first,mahler2019learning,schmidt2018grasping}. This representation allows to reduce the complexity of precise control to simply close the fingers after having reached a particular end-effector pose. However, this requires to define a finite set of pre-grasp shapes which limits the dexterity of a multi-fingered robot hand, e.g.~\cite{kappler2015leveraging,schmidt2018grasping,veres2017modeling,liang2019pointnetgpd,lu2020planning}. Also the resulting grasps are typically power grasps which are great for pick-and-place. However, for assembly or similar tasks, precision grasps are more desirable. A model that outputs contact points per finger basically defines such a precision grasp. Through inverse kinematics, it yields an end-effector pose in $SE(3)$ and a finger configuration that exploits the full dexterity of the hand.  

Of the aforementioned related works,~\cite{varley2015,veres2017modeling,lu2020planning} use a multi-fingered robot hand and go beyond a small set of pre-grasp shapes. \cite{varley2015} suggest to use a CNN that produces heatmaps which indicate suitable locations for fingertip and palm contacts. These contacts are used as a seed for a grasp planner to determine the final grasp pose for the reconstructed object. In follow-up work \cite{varley2017shape}, a CNN completes object shapes from partial point cloud data to provide the grasp planner with a more realistic object model. \cite{veres2017modeling} propose to train a deep conditional variational auto-encoder to predict the contact locations and normals for multi-fingered grasps given an RGB-D image of an object. \cite{lu2020planning} propose a method that directly optimizes grasp configuations of a multi-fingered hand by backpropagating through the network with respect to these parameters. All these methods work for a specific hand while our method not only generalizes over unseen objects but also over novel robotic grippers.


\subsection{Representation Learning for Point Clouds}
Recently, various approaches for processing sparse point clouds have been proposed in related work. \cite{qi2017pointnet} propose PointNet to learn object features for classification and segmentation. PointNet++~\cite{qi2017pointnetPP} applies PointNet hierarchically for better capturing local structures. In our approach, we adopt PointNet~\cite{qi2017pointnet} to extract features of robotic hands and PointNet++ to extract object features. \cite{achlioptas18a} train an auto-encoder network to reconstruct point clouds. The learned representations enable shape editing via linear interpolation inside the embedding space. We train a similar auto-encoder network to learn a representation of various robotic hands.

\section{Technical Approach}
\label{sec: tech}

\begin{figure}[t!]
    \centering
    \includegraphics[width=0.83\linewidth]{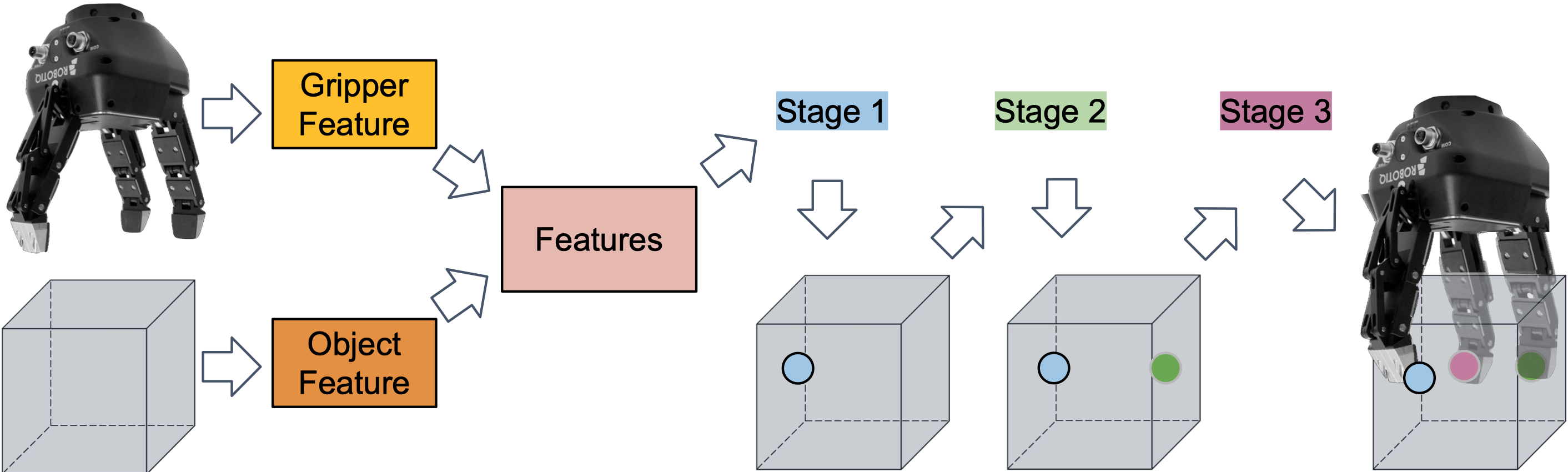}
    \caption{Overview of \emph{UniGrasp} for a 3-fingered gripper. First, we individually map gripper and object features into a lower-dimensional latent space. Their representations are concatenated and fed into our multi-stage \emph{Point Set Selection Network (PSSN)} that generates contact points. The contact points are in force closure and yield a collision-free grasp for the 3-fingered gripper.}
    \label{fig:pipeline}
\end{figure}

Given a point cloud of an object and a URDF of a gripper, we aim to select a set of contact points on the surface of the object such that these contact points satisfy the force closure condition and are reachable by the gripper without collisions as shown in Fig.~\ref{fig:pipeline}. 


If the gripper has $N$ fingers, our model {\em UniGrasp\/} selects a set of $N$ contact points from the object point cloud. In this section, we first discuss how we learn a uniform representation of $N$-fingered robotic hands. This representation can be constructed from a URDF file which describes the kinematics and geometry of the hand. Furthermore, we describe how $N$ contact points are generated on the object surface. We conclude this section by describing the training of the neural network model named \emph{Point Set Selection Network (PSSN)}. We report how we generate a large-scale annotated data set with objects being grasped by a diverse set of two and three-fingered robotic hands in Sec.~\ref{sec: dataset}. Because we have annotated data only for two and three-fingered robotic hands, our trained network model works for two and three-fingered robotic hands. Conceptually, our model could be extended to hands with more fingers if the training data was available.

\subsection{Gripper Representation}
We aim to find a compact representation of the gripper geometry and kinematics as input to {\em UniGrasp\/}.
In this section, we first describe how we encode the geometry of robotic hands. Then we describe how we project a novel robot hand into the learned latent space and construct the input feature (see Fig.~\ref{fig:gripper}).

\subsubsection{Unsupervised Learning of a Gripper Representation}
We use an autoencoder to learn a low-dimensional latent space that encodes the geometry of robotic hands in a specific joint configuration. For training,  we synthesize 20k point clouds of 2000 procedurally generated grippers each in different joint configurations. The generated 2K grippers are two and three-fingered grippers. Their diameter varies from 10 to 30cm. These grippers have prismatic and revolute joints. The number of joints varies from one to twelve.

Each point cloud consists of 2048 points. We use PointNet~\cite{qi2017pointnet} as encoder without the transformation modules. This yields a k-dimensional feature vector that forms the basis for the latent space. The decoder transforms the latent vector using 3 fully connected layers. The first two layers use a ReLU activation function. The last layer produces a 2048$\times$3 output, the reconstructed point cloud of the gripper. We use the Chamfer distance~\cite{fan2017point} to measure the discrepancy between the input and reconstructed point cloud.

\subsubsection{Robotic Hand Representation}
The autoencoder described in the previous section allows us to encode a robot hand in a specific joint configuration. However, we also want to encode the gripper kinematics and range of joints. In the following, we describe how we parse the URDF~\cite{urdf} file and construct a gripper feature for the example of a 3-fingered hand with 3 DoF.
 
 Let us assume, each finger of the hand has one revolute joint. We denote the joints as $\theta_1$, $\theta_2$ and $\theta_3$. Each joint has joint limits. Let $L_i$ and $H_i$ represent the minimum and maximum joint angle. There are $2^3=8$ joint configurations that outline the boundaries of the configuration space of the hand, e.g. $(L1,L2,L3)$ $(H1,L2,L3)$ $(L1,H2,L3)$. We refer to these as boundary configurations. Let $M_i = 0.5 H_i + 0.5 L_i$. Then $(M_1,M_2,M_3)$ describes the joint configuration of the hand where each joint angle takes the mean value of joint limits. We refer to this as central configuration. We generate point clouds sampled on the surface of the gripper model under all boundary and the central configurations. For the above example of a 3DoF gripper, there are 9 point clouds in total. They represent the kinematic and geometric attributes of this specific robotic hand.

\begin{figure}[t!]
    \centering
    \includegraphics[width=0.85\linewidth]{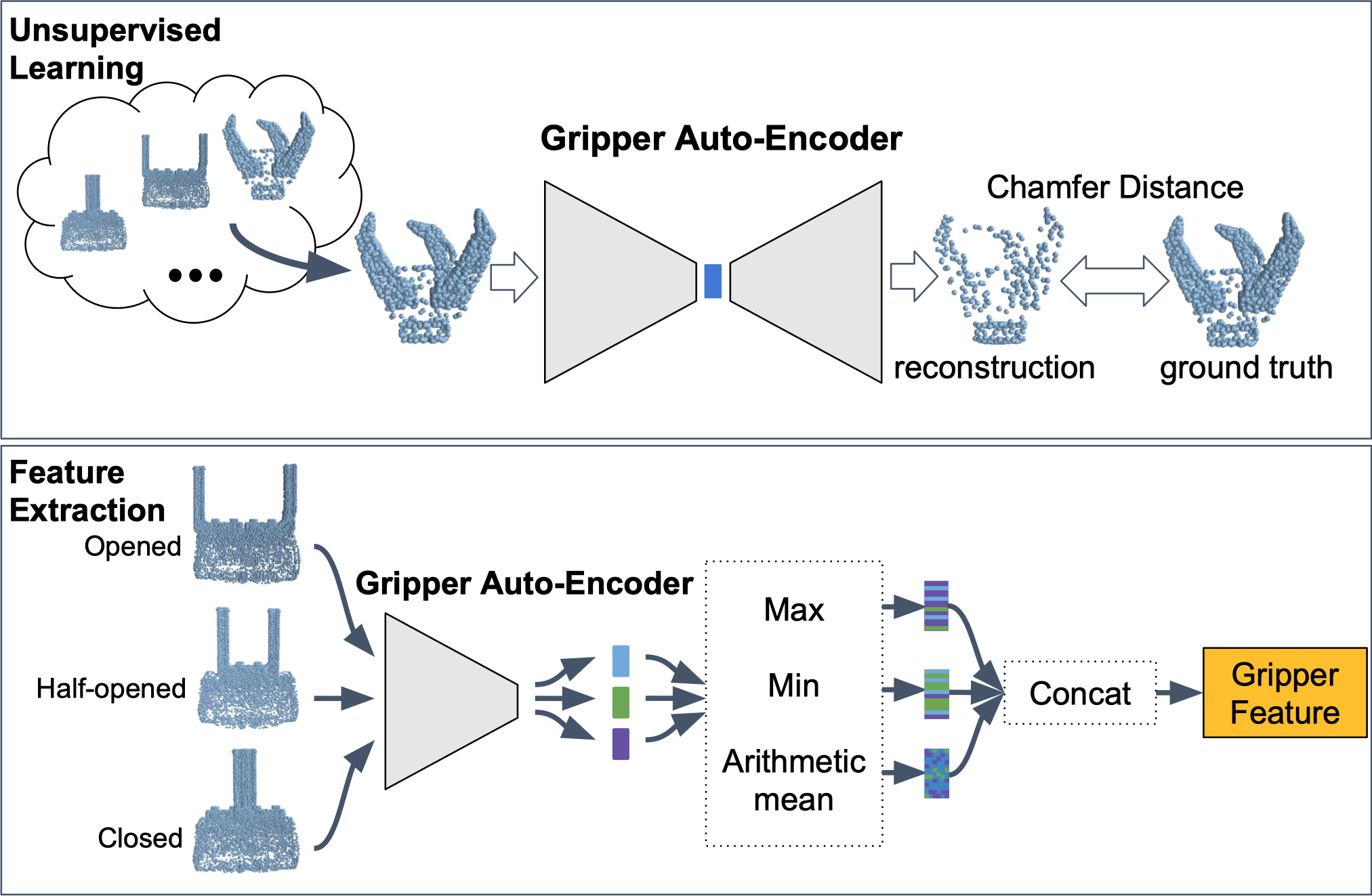}
    \caption{Top: We train an autoencoder to learn a lower-dimensional representation of our grippers and their random configurations by minimizing the Chamfer distance between the reconstructed and the ground truth gripper point clouds. Bottom: The trained encoder is used to generate a feature representation of the geometry and kinematics of the input gripper.}
    \label{fig:gripper}
\end{figure}

 We feed the 9 point clouds into the autoencoder to extract features (see Fig.~\ref{fig:gripper}). Note that robotic hands with a different number of DoF have a different number of boundary configurations. To get a fixed-size feature, we apply three pooling operations (max-, min-, and mean-pooling) among the batch dimension. The output features of those three operations are concatenated to get a final robotic hand feature.

\subsection{Contact Point Set Selection}\label{sec::tech_pssn}

\begin{figure*}[hbt!]
\centering
    \centering
    \includegraphics[width=0.97\linewidth]{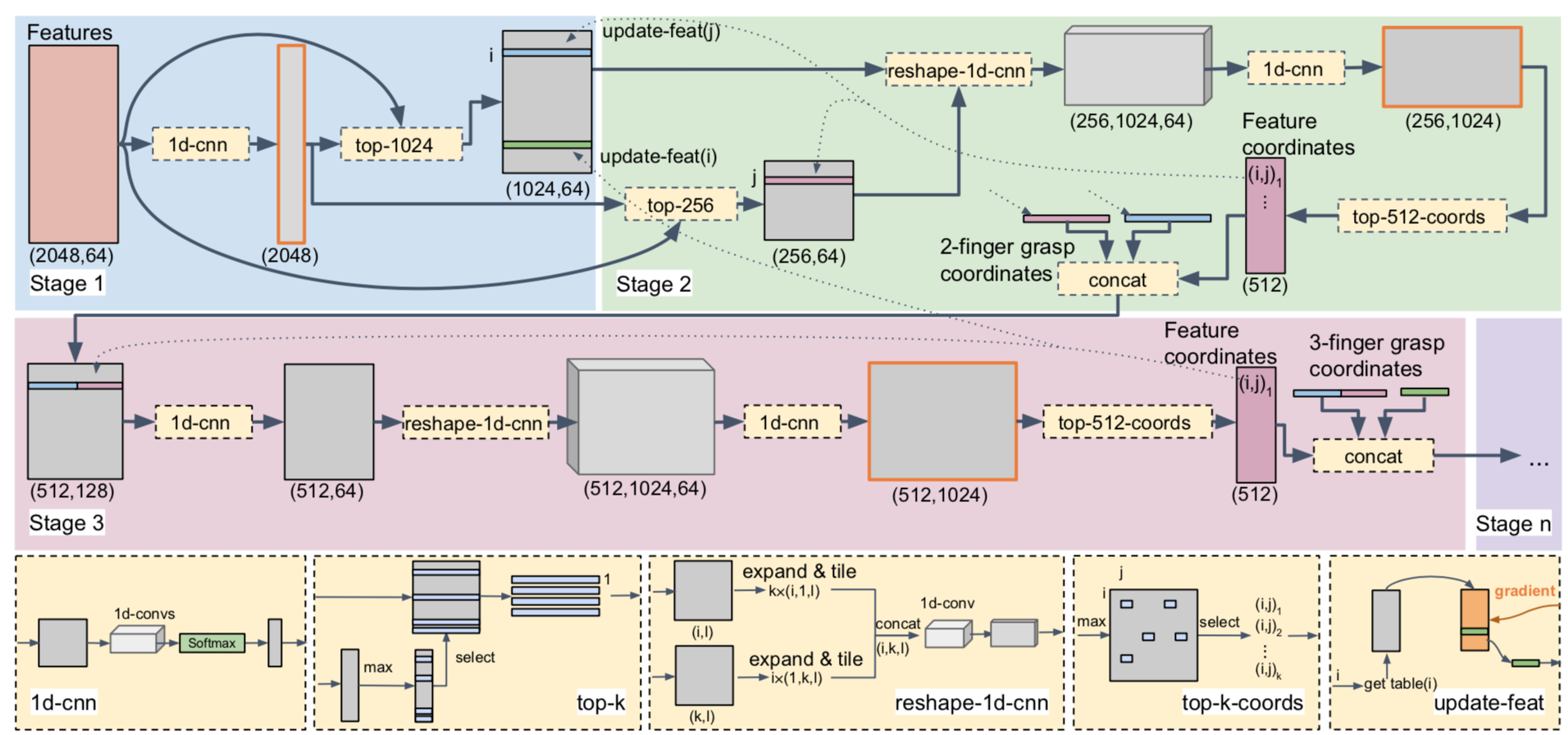}
    \caption{The above scheme describes the multi-stage process shown in Fig. \ref{fig:pipeline} in more detail. Given a feature matrix as input (comprised of gripper and object features), \emph{PSSN} successively expands into $n$ stages to generate the set of contact point coordinates of an $n$-fingered gripper. The two top rows describe the general flow of information from left to right by repeatedly using the modules from the bottom row. Solid lines indicate inputs to modules while dotted lines signify a read (no edge label) or write (\emph{update-feat}) access to previous stage elements. Stage One is primarily used to find a representation of the subset of features that are most promising contact point sets. This representation is used as a template for the following stages (here: Two and Three), which each maintain their own copy of it. These subsequent stages inflate their input before deflating it into feature coordinate vectors (in purple). The coordinates indicate a) which template-copy elements should form the input for the next stage and more importantly b) the indices of data points in the point cloud representing the final grasp locations. Each stage adds one coordinate.}
    \label{fig::model}
\end{figure*}

We use PointNet++~\cite{qi2017pointnetPP} to extract features of the object point cloud represented by $S_0=\{p_i\}_{i=1}^{L}$ where $L$ denotes the number of points. The object feature has a dimension of $L\times64$ and the gripper feature has a dimension of $1\times768$. We repeat the gripper feature along the first dimension by $L$ and change it to be $L\times768$. We then concatenate gripper feature and object features along the second dimension leading to a new dimension of $L\times832$. We apply 1D convolution of the feature of $L\times832$ and the new feature is $L\times64$ denoted as $F_0$ which is the input to the point set selection network.

We aim to select N points from this point cloud such that these N points form force-closure and are reachable by the robotic hand without collisions. The task of training a neural network for point set selection is new and challenging.

Fig.~\ref{fig:pipeline} give a high-level overview of the multi-stage point selection process in \emph{PSSN}. In Stage One, our model first selects one point $p_a$ from the original point cloud $S_0$. Conditioned on the point $p_a$, the network continues to select the second point $p_b$ in Stage Two. If the robotic hand is two-fingered, the point selection procedure ends. The two points $(p_a, p_b)$ are the two contact points to grasp. For a three fingered robotic hand, the third point $p_c$ will be selected during Stage Three. In this paper, we only report results for two and three-fingered hands. For robotic hands with more fingers, our model can conceptually continue to select contact points by adding more stages. In each Stage $n$, the neural network will select the next point conditioned on the previously selected $n-1$ points.
 
For one object and an N-fingered gripper, we denote the list of all valid point sets as $\mathcal{V}=\{(\hat{p}_{i=1}^N)\}$. An item in this list is a set containing N points. A set is said to be valid if it forms force closure and is reachable by the associated robotic hand. However, it is challenging to find these valid point sets among the vast number of possible contact point sets in the object point cloud. To add robustness to the point selection procedure, we adopt an approach akin to beam search \cite{beamsearch} where during each stage the network selects multiple, most promising contact points. The detailed architecture of \emph{PSSN} is shown in Fig.~\ref{fig::model}. In the following, we describe each stage in detail. 

\subsubsection*{Stage One}
We first select $K_1$ points denoted as $S_1$ from the original point cloud $S_0=\{p_i\}_{i=1}^{L}$ as follows. The feature $F_0$
with a shape of $(L,64)$ are fed into a 1d convolution and soft-max layer to calculate the probability of each point being valid. The model then selects the $K_1$ points with the highest scores and gathers the corresponding features denoted as $F_1$ with a dimension of $(K_1,64)$.

\subsubsection*{Stage Two} 
This stage aims to select points from $S_0$ that conditioned on the points in $S_1$ yield valid point sets.
As a first step, Stage Two selects $K_2$ points from $S_0$ in the same way as Stage One. This set of points is referred to as $S_2$. The network collects the corresponding point features in $F_2$ which is of dimension $(K_2,64)$. Then, feature maps $F_1$ and $F_2$ are copied and put into a reshape-1d-cnn layer visualized in Fig.~\ref{fig::model}. The network performs a pairwise copy and concatenation to produce a combined feature map $F_3$ of dimension $(K_2,K_1,64)$. The feature map is then sent to 1d convolution and soft-max layer to get the final score matrix of dimension $(K_2,K_1)$. We collect the top $K_3$ elements denoted as $S_3$. For two fingered robotic hands, these two points are the contact points predicted by \emph{PSSN}.

\subsubsection*{Stage Three}
In this stage, the network selects a third contact point. For this purpose, the network accesses the features of each element in the set $S_3$. Each index in the set $S_3$ refers to two previously selected points of Stage One and Stage Two. The network copies the corresponding features and concatenates them. The new feature map has a dimension of $(K_3,128)$ and is fed into a 1d convolutional layer with an output of dimension $(K_3,64)$ denoted as $F_4$. We select the third point index from the already reduced set $S_1$. For this purpose, the network simultaneously gathers the feature map $F_4$ with dimension $(K_3, 64)$ and feature map $F_1$ with dimension $(K_1, 64)$. These two feature maps are fed into a reshape-1d-cnn layer to get a pairwise concatenation with a dimension of $(K_3,K_1,64)$ denoted as $F_5$. The feature map $F_5$ is input to a 1d convolution and soft-max to calculate the score matrix of dimension $(K_3,K_1)$. Each element $(u,v)$ in this matrix corresponds to the $u_{\text{th}}$ element in $S_3$ and the $v_{\text{th}}$ point in set $S_4$. Note that the $u_{\text{th}}$ element in the set $S_3$ refers to two points selected in the previous two stages. Based on the score, the network determines the $K_4$ top contact points sets. These form the candidate grasps for the three-fingered robotic hand. 

In practice, we find it is easier to train the \emph{PSSN} when rejecting a large number of invalid point sets. The following heuristics reduce the number of candidate triplets from billions to millions. As a first step, \emph{PSSN} predicts a normal per point in this point cloud. We denote these predicted normal vectors $\{\hat{pn}_i\}_{i=1}^{L}$. Then, we leverage simple heuristics~\cite{borst2003grasping} to reject invalid point sets based on point positions and the predicted point normals.  
For two-fingered robotic hands, the angle between the two contact point normals needs to be larger than $120^\circ$ to be in force closure~\cite{nguyen1988constructing}. For three-fingered robotic hands, three contact points form a triangle. 
We set a constraint that each side of the triangle needs to be larger than 1 cm to prevent our model from selecting points which are too close to each other. In addition, we prefer contact points that form a more regular triangle and we set another two constraints. The maximum internal angle of the triangle needs to be less than $120^\circ$. For each point, the angles between its normal vector and two incident edge directions are less than $90^\circ$. It indicates that the point normal is pointing outwards of the triangle.  Generally three contact points in force closure fulfill those constraints above. 

\subsection{Loss Function and Optimization}
Given the ground truth point normals $\{pn_i\}_{i=1}^L$, the loss of predicted point normals is defined to be
\begin{equation*}
    L_{pn} = -\sum_{i=1}^{L}(\hat{pn}_i \boldsymbol{\cdot} pn_i)
\end{equation*}
where $\boldsymbol{\cdot}$ represents the dot product.

\citep{kappler2016optimizing} found that formulating grasp success prediction as a ranking problem yields higher accuracy than formulating it as a binary classification problem. We therefore formulate point selection at every stage as a learning-to-rank~\cite{pasumarthi2019tf} problem. 
For an object point cloud and a gripper, every point set which is in force closure and reachable by this specific robot hand is annotated as positive. All other point sets are labeled negative. Therefore, each point set has a binary label $y$. 
In Stage n (where n $\leq$ N), the network predicts a list of $K_n$ point sets and their corresponding probability $\hat{y}$ of whether this set is valid or not. Based on their predicted probabilities, we have a ranked list of these point sets. We use a variant of the ListNet~\cite{cao2007learning} loss to increase the $\hat{y}$ of point sets with positive labels.
\begin{equation*}
    \mathcal{L}_n = -\sum_{j=1}^{K_n} y_j \log(\frac{\exp(\hat{y}_j)}{\sum_{j=1}^{K_n} \exp(\hat{y}_j)})
\end{equation*}

We use the Adam optimizer~\cite{kingma2014adam} for training the network, set the learning rate to $10^{-4}$, and split the data into 80/20 for training and test sets. We train the neural network stage by stage. First, we only train Stage One. Only the loss of Stage One is computed and the gradients are back-propagated to the weights of the layers in Stage Two.  After training Stage One, we fix the weights in Stage One and continue to train the layers in Stage Two. Only the loss of Stage Two is computed. The training procedure continues until Stage N where in this paper $N=2,3$.
\section{Grasp Dataset Generation}\label{sec: dataset}
\begin{figure}[t!]
    \centering
    \includegraphics[width=0.85\linewidth]{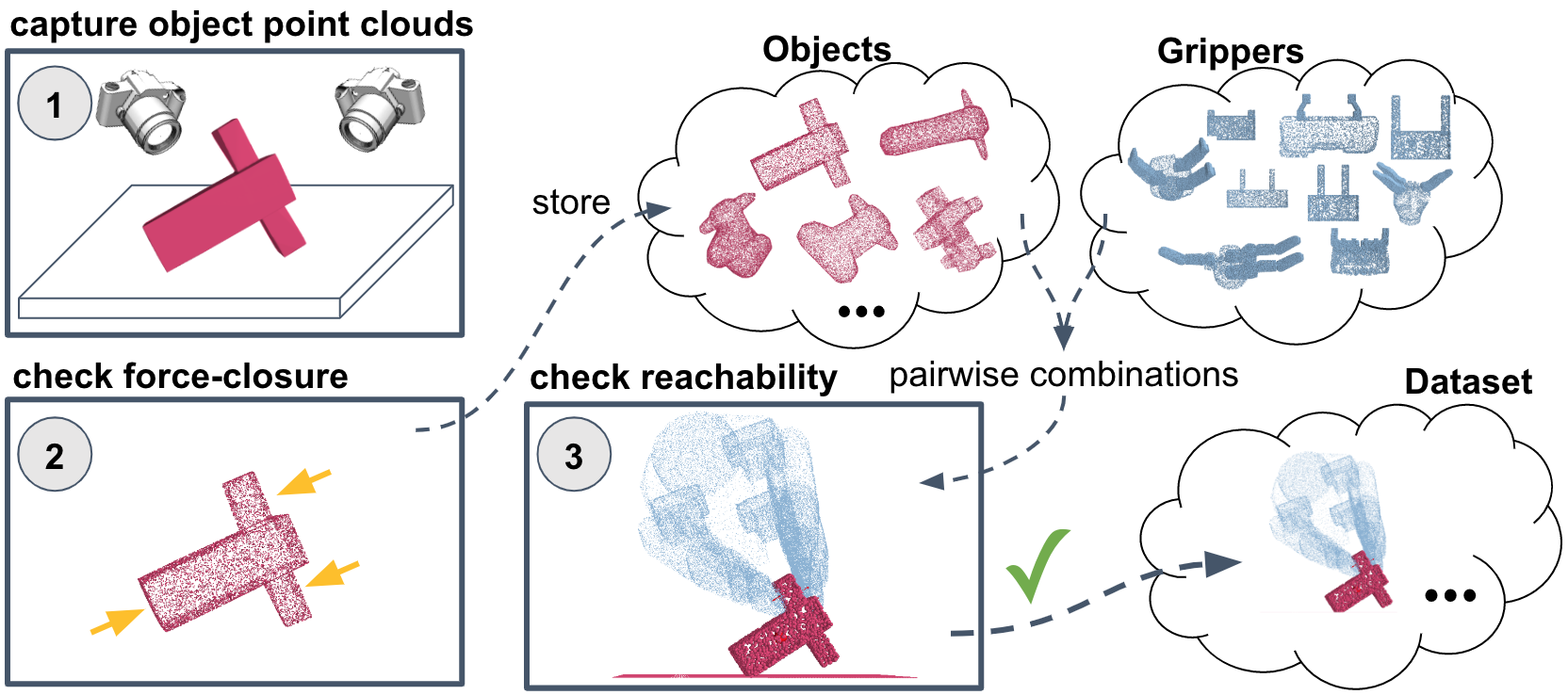}
    \caption{Overview of our dataset generation process. Point clouds of different Bullet \cite{coumans2019} object meshes were generated based on images rendered from 8 different viewpoints around the objects. All point sets passing a force-closure check were then added to the list of valid points. If these point sets are also reachable and collision-free by a specific gripper, then it is added to the final dataset used to train \emph{UniGrasp}.} 
    \label{fig:data}
\end{figure}

To train the \emph{UniGrasp} model, we require data that consists of object point clouds annotated with sets of contact points that are in force closure and reachable by specific grippers. We generate this data set in simulation as commonly done for other data-driven approaches~\cite{kappler2015leveraging,mahler2016dex,mahler2017dex,mahler2019learning}. To construct this dataset, we select 1000 object models that are available in Bullet~\cite{coumans2019} and scale each object up to five different sizes to yield 3275 object instances. We use 12 different robotic hands. Nine of these hands are two-fingered robotic grippers and three of them are three-fingered~\footnote{We use commercially available grippers such as the Sawyer \cite{sawyer} and Panda \cite{panda} parallel-yaw gripper, Robotiq 3-Finger \cite{robotiq3f}, Kinova KG-3 \cite{kinova}, and Barrett BH8-282 gripper \cite{barrett}}. The data generation process is visualized in Fig.~\ref{fig:data}.

We place the object on a horizontal plane and render the object from eight viewpoints to generate RGB-D images. We reconstruct a point cloud from those RGB-D images and down-sample them to 2048 points. We find 2048 points are dense enough to represent the geometry of the object given GPU memory constraints.  

Given a robotic hand with $N$ fingers (N= 2,3), we generate all possible sets of N points that are in force closure and reachable by the given robotic hand. For an object represented by a point cloud of 2048 points and a three-fingered robotic hand, the total number of combinations is $\binom{2048}{3}\approx 1.4\times10^{9}$. We assume a friction coefficient of $0.5$ for two-fingered grippers and of $0.65$ for three-fingered grippers and use a polygonal approximation of the friction cone with 16 faces. We first utilize the heuristics described in Section~\ref{sec::tech_pssn} to reject large amounts of point sets. We use FastGrasp~\cite{pokorny2013c} to evaluate the point sets and compute the grasp quality score $Q^{-}_{l}$. 
We label a point set to be in force closure for two-fingered grippers, if $Q^{-}_{l} > 0$ and a point set to be in force closure for three-fingered grippers, if $Q^{-}_{l} > 0.0001$.

If a set of $N$ points is in force closure, we use inverse kinematics for each $N$-fingered hand to determine (a) whether the points are reachable and (b) whether the grasp is collision-free - both specific to the hand. 
All sets of contact points that are in force closure, reachable and collision-free are labeled as positive for a specific gripper and negative otherwise. 

\section{Experiments}\label{sec: exp}
In this section, we extensively evaluate the proposed grasp synthesis method in terms of the accuracy of contact point selection for various known N-fingered robotic hands both in simulation and in real world experiments. We demonstrate the validity of the learned gripper embedding space and show that our model can produce valid contact points for different novel robotic hands, both in simulation and the real world experiments.

\subsection{Robotic Hand Presentation Learning}
\label{sec:handrepresentationlearning}
For training the autoencoder to represent the gripper, we split the gripper point clouds described in Sec.\ref{sec: tech} into training and test set with a ratio of nine to one. We use five consecutive layers of 1D convolution in PointNet~\cite{qi2017pointnet}. 
The feature channel dimensions of each convolution layers are (64,64,128,128,256,256) starting from input layers. The final feature embedding space has a dimension of 256. The training and test Chamfer Distance losses~\cite{fan2017point} are $4.87\times10^{-5}$ and $4.44\times10^{-5}$ respectively.

\begin{figure}[t]
    \centering
    \includegraphics[width=0.82\linewidth]{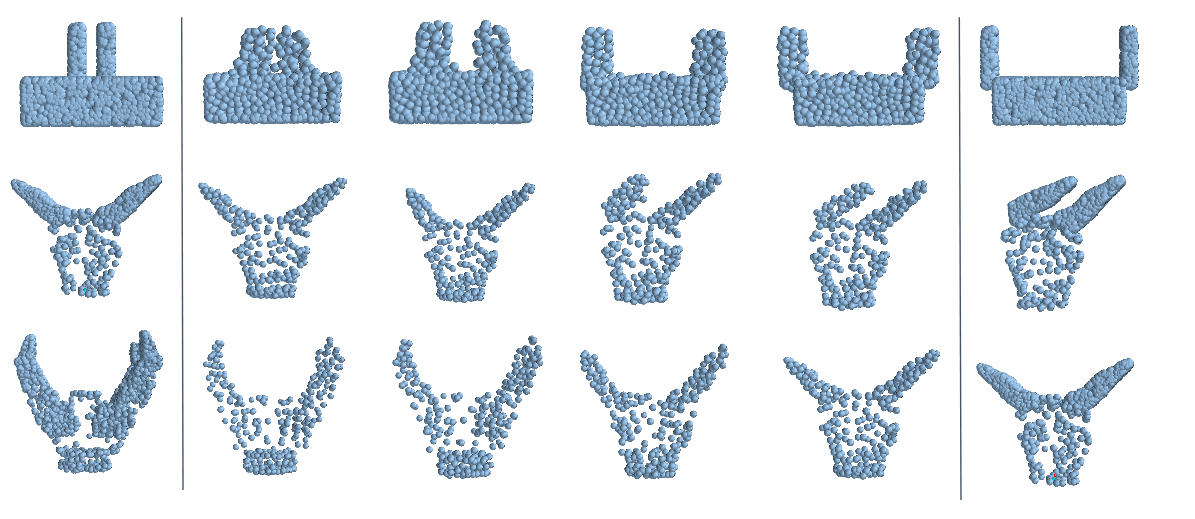}
    \caption{Interpolation in Gripper Feature Space. The top row indicates prismatic joint movement, the middle row represents revolute joint movement. The bottom row shows the geometry changes between two types of 3-fingered robotic hands.}
    \label{fig:interp}
\end{figure}

To demonstrate that the learned representation is able to capture robotic hand attributes like prismatic and revolute joint movement as well as variety in geometry, we show the linear interpolation ability between two points in the latent space. We take two point clouds $Pc_1$ $Pc_2$ and feed them into the autoencoder. We extract the corresponding features from the encoder denoted as $F_1$ and $F_2$. We generate new features $F_3 = 0.8F1 + 0.2F2$ and $F_4 =0.2F1 + 0.8F2$ by interpolation between $F_1$ and $F_2$. $F_1-F_4$ features are put into the decoder to reconstruct point clouds. The results are shown in Fig.~\ref{fig:interp} and indicate the meaningful interpolation not only in configuration space but also in shape space.

\subsection{Grasp Point Set Selection Performance}

\begin{figure*}[hbt]
\centering
\footnotesize
\resizebox{\textwidth}{!}{
\begin{tabular}{l||cc|cc|cc|cc|cc|cc|c}
\hline
\hline
\multirow{2}{*}{Accuracy} & \multicolumn{2}{c|}{Stage One Train} & \multicolumn{2}{c|}{Stage One Test} & \multicolumn{2}{c|}{Stage Two Train} & \multicolumn{2}{c|}{Stage Two Test} &\multicolumn{2}{c|}{Stage Three Train} & \multicolumn{2}{c|}{Stage Three Test} &  \multicolumn{1}{c}{Neighbor}\\
        & \multicolumn{1}{l}{Top1} & \multicolumn{1}{l|}{Top10} & \multicolumn{1}{l}{Top1} & \multicolumn{1}{l|}{Top10} &
        \multicolumn{1}{l}{Top1} & \multicolumn{1}{l|}{Top10} &
        \multicolumn{1}{l}{Top1} & \multicolumn{1}{l|}{Top10} & 
        \multicolumn{1}{l}{Top1} & \multicolumn{1}{l|}{Top10} & \multicolumn{1}{l}{Top1} & \multicolumn{1}{l|}{Top10} &
       \multicolumn{1}{l}{}  \\ \hline
T Sawyer  & 99.8 & 100.0  & 99.6 & 100.0 & 93.6 &  97.6 & 92.8 & 96.8 &- & - & - & - & 53\\
U Robotiq-2F & - & - & 93.8 & 100.0  & - & -  & 75.8 & 91.7 & -& - & - &- &  37\\
U Sawyer & - & - & 97.4 & 100.0  & - & -  & 86.6 & 95.4 & -& - & - &- &  43\\
B Sawyer & - & - & - & -  & - & -  & 66.2 & 88.8 & -& - & - &- &  -\\
T Franka & 99.0  & 99.8 & 98.7 & 100.0 & 84.7 &  95.6 & 83.7 & 91.0 & -& -&- &- & 43\\ 
T Kinova-3F & 96.6 & 99.7 & 96.8 & 99.8 & 91.6 & 96.3 & 91.6 & 96.3 & 76.3 & 89.5 & 77.1 & 90.7 & 26\\
T Barrett & 98.4 &  100.0 & 98.7  & 99.8 & 95.6 & 98.3 & 97.1 & 98.9 & 90.6 & 96.4 & 89.6 & 96.3 & 40\\  
T Robotiq-3F & 98.2 & 99.8 & 98.4 & 99.6 & 95.6 & 98.3 & 96.3 & 98.4 & 86.6 & 94.2 & 86.9 & 95.1 & 36\\
B Robotiq-3F & - & - & - & -  & - & -  & - & - & -& - & 20.9 & 41.0 &  - \\
\hline
\end{tabular}}
\caption{Summary of our grasping evaluation in simulation for two-fingered and three-fingered grippers. We report prediction accuracies of \emph{PSSN} at different stages for different grippers. We distinguish between two different evaluation strategies and denote them by \emph{T} and \emph{U} respectively. A gripper prefixed with \emph{T} (e.g. \emph{T Sawyer}) indicates the gripper was trained and evaluated on train-test-splits dataset while both splits contained samples of that gripper (among other grippers). In contrast, \emph{U}-prefixed grippers denote that samples of the gripper were not used during training. The gripper prefixed with \emph{B} presents the grasp performance of that gripper using baseline method. The column Neighbor shows the percentage of valid grasps among all the existing point sets in a neighborhood of the selected grasp.} 
\label{fig::result}
\end{figure*}

In simulation, \emph{PSSN} selects $K_1=1024$ in Stage One, $K_2=1024$ and $K_3=1024$ in Stage Two and $K_4=1024$ in Stage Three. We run six experiments to demonstrate the strength of our proposed method in predicting valid contact points for up to three-fingered robot hands. 

\subsubsection{Evaluation Metrics}
We adopt two evaluation metrics to reflect prediction accuracy. \textbf{Top1} refers to the prediction accuracy of the highest ranked point to be within 5mm of a valid point set $\mathcal{V}$, i.e. of a point set that forms a force closure grasp, is reachable and collision-free by the input hand. \textbf{Top10} refers to the percentage of test cases where at least one valid point set of the top 10 predicted ones is within 5mm of a valid grasp. We allow the 5mm relaxation for two reasons. a) There is always annotation noise e.g. in the point normal or force closure estimation. b) In the real world, fingertips often have a width of 10mm. With a 5mm tolerance the fingertip still covers the valid contact point. A similar relaxation is also used in previous work~\cite{le2010learning}. 

\textbf{Neighbor} reports the ratio of valid point sets over all possible point sets within a local neighborhood of our prediction. We set the radius of this local neighborhood around each contact point to be 5mm. A high number shows that there are many valid points around a predicted contact point. This is helpful during grasp acquisition in the real world where noise in the point clouds and in robot actuation may lead to alignment errors between the robot fingers and contact points. 

\subsubsection{Baseline}
There is no previous work that has proposed a model for grasping novel objects with novel grippers. Therefore, we propose the following baseline that is close to our approach on data generation but is not based on learning. Given a partial point cloud of an object, we apply the same aforementioned heuristics to reject contact point sets that are unlikely to be feasible grasps. We run FastGrasp~\cite{pokorny2013c} to evaluate the remaining point sets. FastGrasp requires as input the object's center of mass and the surface normals at the contact points. As we do not assume access to these properties in UniGrasp, we approximate them for FastGrasp. The object's center of mass is the average position of all points in the point cloud. To estimate the normal of each point, we compute the covariance matrix of the nearest 30 points. We adopt the eigenvector with the smallest eigenvalue as normal. The normals are oriented towards the estimated center. 

Given this information, we iterate over contact point sets with estimated point normals and center of mass until FastGrasp returns a point set that is in force closure. If this point set is also reachable and collision-free, we consider the grasp successful. If not, we continue iterating over candidate point sets. If FastGrasp fails to return a  point set which has positive label in the dataset within 10 second, we consider the object grasping as a failure. 

\subsubsection{Results}
To test our model's generalization ability to \textbf{novel objects}, we train a neural network on the training dataset and report its performance on the test dataset. The results are given in Fig.~\ref{fig::result}. We parse the URDF of various grippers and generate their features as described in Sec.~\ref{sec: tech}. During each training stage, our model selects point sets and ranks them. For example, given a two-fingered robotic hand, the network has two prediction stages. 

For two-fingered grippers, our model achieve 92.8\% and 83.7\% \textbf{Top1} accuracies for Saywer and Franka, respectively. We also test the grasping performance for three-fingered robotic hands. We train our point set selection network on the objects with annotation of Kinova, Robotiq and Barrett hand, respectively. Our model achieves 77.1\%, 86.9\% and 89.6\% \textbf{Top1} test accuracy. 

We run the baseline using the two-fingered Sawyer gripper and the three-fingered Robotiq-3F gripper and compare it with UniGrasp on the same hands. Our UniGrasp method achieves 86.9\% and 89.6\% average grasp success rates, respectively. The baseline only achieves 66.2\% and 20.9\% average grasp success rates. Therefore, UniGrasp significantly improves over the baseline.

To test our model's generalization ability to \textbf{novel robotic hands}, we first train the neural network model on the dataset annotated with all two-fingered grippers except the Sawyer and Robotiq-2F. In the test stage, we first parse the URDF file of these two new grippers and extract corresponding gripper features. We then give these features as input to PSSN and evaluate the grasp performance on the test dataset annotated for each gripper. Our model achieves an accuracy of 86.6\% and 75.8\% on the \textbf{Top1} prediction. There are on average 43\% and 37\% valid points that are \textbf{Neighbors} of the predicted contact point. This shows that our model clusters valid contact point sets in regions on the object surface. This is beneficial for real-world grasp executions under perception and actuation noise as for example discussed in~\cite{roa2009computation}.

\begin{figure}[hbt!]
    \centering
    \includegraphics[width=0.8\linewidth]{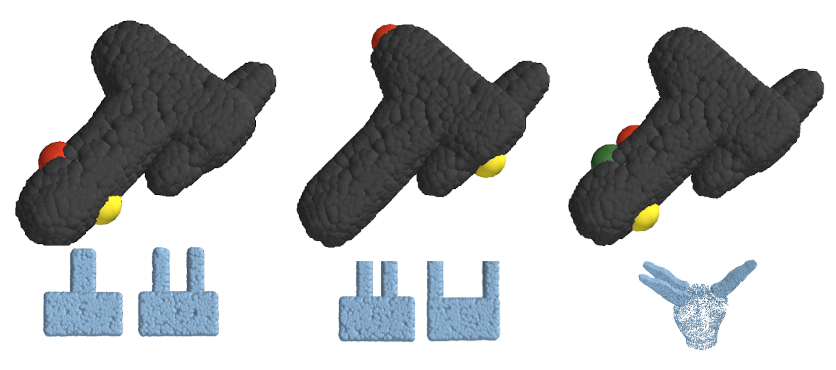}
\vspace{-4mm}
    \caption{Examples of contact point selection results on one object with two-fingered grippers and Kinova-kg3. The black point cloud represents the object in top view. Red, yellow and green spheres indicate the contact point predicted by our \emph{PSSN} in Stage One, Two and Three respectively. Blue points represent the gripper. For two-fingered grippers, left gripper represents the full-closed status and right gripper represents the full-open status. Our method generates contact point sets that are adapted to robotic hand attributes.}
    \label{fig:neighbor}
\end{figure}

Fig.~\ref{fig:neighbor} illustrates that our model generates distinct contact point sets on the same example object for various robotic hands. This qualitatively shows that our model generates contact points that are specific to the robotic hand attributes. We also conducted an experiment that quantitatively analysis how much the gripper features matter in generating valid contact points. For this, we denote the left two-fingered gripper G1 and the right two-fingered gripper G2 in Fig.~\ref{fig:neighbor}. We send the gripper feature of G1 and evaluate the predicted results using annotated data of G2 (denoted as G1/G2). We run four experiments which are G1/G1, G1/G2, G2/G1, G2/G2. In Fig.~\ref{fig::res}, we show that the grasp success rate drops by 30-20\% when the test hand does not match the input hand. This demonstrates that PSSN leverages the gripper embedding to generate contact points which are specific to the input gripper. 

\begin{figure}[t!]
\centering
\scriptsize
\resizebox{\linewidth}{!}{
\begin{tabular}{l||cc|cc|c}
\hline
\hline
\multirow{2}{*}{Accuracy} & \multicolumn{2}{c|}{Stage One Test} & \multicolumn{2}{c|}{Stage Two Test} &  \multicolumn{1}{c}{Neighbor}\\
        & \multicolumn{1}{l}{Top1} & \multicolumn{1}{l|}{Top10} & \multicolumn{1}{l}{Top1} & \multicolumn{1}{l|}{Top10} &
       \multicolumn{1}{l}{}  \\ \hline
G1/G2 & 82.3 & 95.2 & 26.6 & 30.7 & 8.8\\
G1/G1 & 89.9 & 98.0 & 55.8 & 60.8 & 27.5\\
G2/G1 & 70.6 & 85.9 & 54.3 & 61.7 & 32\\ 
G2/G2 & 77.1 & 92.6 & 73.8 & 81.2 & 40\\
\hline
\end{tabular}}
\caption{ \emph{G1} and \emph{G2} denote the left and middle gripper shown in Fig.~\ref{fig:neighbor}, respectively. G1/G2 means the input gripper is G1 and the test ground truth data is G2. G1/G1 means the input gripper is G1 and the test ground truth data is G1.} 
\label{fig::res}
\end{figure}

\subsection{Real World Experiments}
We evaluate the performance of our model in real world experiments on 22 objects which are shown in the corresponding video. The object set includes eight objects from Dexnet2.0~\cite{mahler2017dex}, three objects from the YCB object data set~\cite{7254318} and two deformable objects. We use a laptop with Intel Core i9 CPU and Nvidia 1070 GPU to run our neural network model. Our hand-eye camera setup uses an RGB-D camera to capture the depth images of a given object. We use a single depth image for each grasping experiment. The depth image is inpainted~\cite{telea2004image} using OpenCV to remove invalid values. Utilizing the camera matrix, we can reconstruct a 3D point cloud. We feed the gripper description and object point cloud into our \emph{UniGrasp} model. \emph{UniGrasp} outputs each finger's contact point. We solve the inverse kinematics for the robotic gripper using RBDL \cite{RBDL-Felis2016} which takes around 0.4 milliseconds, and command the robot to approach the object using a Cartesian space controller. We close the fingers and lift the object. PSSN simultaneously selects 256 point sets. We start to go through this list from top1 and use an inverse kinematics solver to compute the joint configuration for reaching the point set. Although UniGraps is trained to output reachable contact points, there is no guarantee. However, if the inverse kinematics solver does not generate a solution or there is a collision, we can select the next best set of contact points in the output list. However, this did not occur in our real-world experiments.

We apply our model in the real world using the Kinova KG-3 and Robotiq-3F grippers that our model was trained on. To test the generalization of our model over grippers, we remove one finger of Kinova KG-3 in the URDF file to create a novel two-fingered gripper. We also apply UniGrasp to the Schunk SVH five-fingered robotic hand~\cite{schunk}. Our \emph{UniGrasp} model produces three contact points for the thumb, index and ring finger. After solving inverse kinematics for those three fingers, we control the five fingers by having the middle finger mimic the index and the pinkie mimic the ring finger. Finally, we also apply our method to the Allegro hand (from Wonik Robotics) using the thumb, index and middle finger, and we compare our method to the baseline in the case of the Allegro hand. 

The results are shown in Fig.~\ref{fig::real_exp}. Our method achieves 92\% and 95\% successful grasps for Kinova-KG3 and Robotiq-3F grippers. It also generates 93\% for the novel two-fingered gripper. Although in our experiment the thumb of the Schunk hand was unfortunately broken, our model achieves 83\% successful grasps. For the Allegro hand, we achieve a 90\% success rate while the baseline only achieves 40\% of successful grasps.

\begin{figure}[t!]
\centering
\footnotesize
\begin{tabular}{c|ccc}
\hline
\hline
& Trials & SuccessRate(\%) & RunTime(s) \\
\hline
T Kinova-KG3 & 64 & 92 & 0.13\\
T Robotiq-3F & 65 & 95 & 0.20\\
U Kinova-2F& 60 & 93 & 0.06\\
U Schunk SVH & 60 & 83 & 0.21\\
U Allegro Hand& 60 & 90 & 0.22 \\
B Allegro Hand& 60 & 40 & 4.87\\
\hline
\end{tabular}
\caption{Summary of our grasping evaluation in the real world for various grippers. The prefix \emph{T} and \emph{U} are the same as in Fig.~\ref{fig::result}. The prefix \emph{B} represents the baseline described in Sec.~\ref{sec: exp}. RunTime represents computation time spent on \emph{PSSN} or the average time spent on heuristics+FastGrasp for the baseline.}
\label{fig::real_exp}
\end{figure}

\section{Conclusion}\label{sec:conclusion}
We present a novel data-driven grasp synthesis approach named \emph{UniGrasp} that generates valid grasps for a wide range of grippers from two-fingered parallel-yaw grippers to articulated multi-fingered grippers. \emph{UniGrasp} takes point clouds of the object and the URDF of robotic hand as inputs. The outputs are the contact points on the surface of objects. We show in quantitative experiments that our method predicts over 90\% valid contact points for various known two- and three-fingered grippers in Top10 predictions and over 90\% valid grasps in real world experiments. Our model also generates 93\%, 90\%, and 83\% successful grasps for novel two-fingered, four-fingered and five-fingered robotic hands in real-world experiments. In future work, we aim to extend UniGrasp to n-fingered hands where $n>3$. For this, we will also reconsider the heuristics used for rejecting contatc point sets.


{\scriptsize
\bibliographystyle{IEEEtranN}
\bibliography{references}
}
\end{document}